# Benchmarking GPT-5 for biomedical natural language processing

**Yu Hou[1], Zaifu Zhan[2], Min Zeng[1], Yifan Wu[1], Shuang Zhou[1] and Rui Zhang[1]**

**[1]Division of Computational Health Sciences, University of Minnesota, Minneapolis, Minnesota, USA.**

**[2]Department of Electrical and Computer Engineering, University of Minnesota, Minneapolis, MN, USA.**

## Abstract

**Background:** Biomedical literature and clinical text pose heterogeneous demands on natural-language systems, ranging from extraction, document synthesis, and multi-step clinical reasoning. Recent large language models (LLMs) have narrowed the gap with task-specific systems; however, it remains unclear how much frontier general models advance accuracy, cost efficiency, and deployment readiness across diverse BioNLP settings.

**Methods:** We extend a standardized benchmark to evaluate GPT-5 and GPT-4o under zero-, one-, and five-shot prompting on five core BioNLP tasks—named entity recognition (NER), relation extraction (RE), multi-label document classification, summarization, and simplification—and an expansion of nine biomedical QA datasets covering knowledge QA, clinical reasoning, and multimodal visual QA. We use unified prompt templates, fixed decoding parameters, and a batch inference pipeline for core tasks, while QA uses the standard API configuration. For every model–task pair we log input/output tokens, compute token-normalized costs under official pricing, and record end-to-end latency. Previously reported GPT-4 and GPT-3.5 results are included for longitudinal comparison.

**Results:** GPT-5 outperformed GPT-4o on the QA expansion with the biggest gains on reasoning-centric datasets like MedXpertQA and DiagnosisArena and consistent advantages on multimodal QA. Despite disease NER and summarization remaining below domain-tuned baselines, GPT-5 improved chemical NER and ChemProt in core tasks. GPT-5 had a similar median latency and lower total cost than GPT-4o in batch mode and cost-per-correct-answer on standard QA despite longer outputs. Fine-grained analyses show that diagnosis, treatment, and reasoning subtypes improve, but boundary-sensitive extraction and evidence-dense summarization remain challenging.

**Conclusions:** With a unified and reproducible protocol, GPT-5 gets close to deployment-ready performance for biomedical QA while also offering good cost-performance trade-offs. Our results support a tiered prompting strategy, direct prompting for extensive or cost-sensitive pipelines, and chain-of-thought scaffolds for analytically intricate or high-stakes situations. We advocate for hybrid solutions for span-level extraction and summarization, where accuracy and fidelity are critical. The benchmark and the cost/latency accounting that accompany it provide useful information for selecting models and designing systems in biomedical NLP.

## Introduction

The rapid growth of biomedical literature, which now comprises more than 38 million PubMed-indexed records and adds over one million articles annually, makes it challenging to curate knowledge, search for literature, and discover new information[1–3]. Specifically, the COVID-19 pandemic led to an unparalleled increase in biomedical publications. By the end of 2024, PubMed had indexed more than 390,000 COVID-19 articles[4,5]. This made it even harder to stay up to date and find useful information. As the amount of information grows, it is becoming increasingly difficult for human experts to sort through it all. This underscores the need for robust biomedical natural language processing (BioNLP) tools even more[6–8]. Conventional BioNLP techniques rely on supervised models that are meticulously adjusted using task-specific datasets; however, these methods require considerable manual annotation and face challenges in generalizing beyond their limited training data[9–12]. Recent progress in large language models (LLMs), particularly transformer-based systems with hundreds of billions of parameters, holds promise as a new approach to doing things[13]. Models such as GPT-3.5 and GPT-4 have demonstrated extensive linguistic comprehension and reasoning abilities, including in medical contexts[14,15].

A systematic evaluation of LLMs across diverse biomedical NLP tasks revealed a complex landscape. Chen et al. performed an extensive evaluation of GPT-3.5, GPT-4, and LLaMA2 (13B) across 12 standard datasets encompassing six task categories[16]. Their results showed that GPT-4's performance in zero-shot and few-shot tasks was impressive, often coming close to or surpassing the best results in medical question answering and performing well in classification tasks. In most tasks that involved extracting information from text or generating new text, it was still behind fine-tuned models. Another open-source evaluation[17] ound that DeepSeek models perform well on NER and text classification, but they still struggle with relation and event extraction due to the trade-off between precision and recall. By keeping the evaluation method, using the same prompt templates and examples available to the public, new models can be tested in the same way to assess the progress made. In a field that moves quickly, this standardized benchmarking is very important. LLMs are frequently updated, and new models are released, so it is necessary to make "apples-to-apples" comparisons for research that can be replicated and to help people choose the best model. Since the first benchmark was completed in early 2025, biomedical NLP has seen two big improvements in LLM. First, OpenAI released GPT-4 Omni (GPT-4o) in May 2024. This was an improved version of GPT-4 that could really use multiple modes and tools[18]. Second, the GPT-5 model was officially released in August 2025. This was the next generation in the GPT family. OpenAI hasn't shared all the details about the architecture of GPT-5, but they say it has better reasoning skills, supports more modes of input, and is easier to use than GPT-4 and GPT-4o.

In this study, we extend the prior BioNLP benchmark to include both GPT-5 and GPT-4o, evaluated alongside previously reported results for GPT-4, GPT-3.5, and LLaMA-2-13B across five text-based task families originally benchmarked by Chen et al.[16]: named entity recognition (NER), relation extraction (RE), multi-label document classification, text summarization, and text simplification. Beyond these foundational tasks, we introduce an extensive suite of biomedical question-answering (QA) and reasoning datasets, encompassing knowledge-based QA, clinical reasoning QA, and multimodal visual QA. This expansion enables a more comprehensive assessment of LLM performance in diagnostic reasoning, textual understanding, and visual-semantic integration within realistic biomedical contexts. In addition to task-specific primary metrics under zero-, one-, and five-shot settings, we record token-level input/output statistics and inference latency to quantify cost and efficiency. For selected QA datasets, we further conduct fine-grained analyses by question or content type—such as subject distributions in MedMCQA[19], question types in VQA-RAD[20], and subtype/modality/content-type breakdowns in SLAKE[21]—to examine performance consistency across reasoning dimensions. The goals of this work are fourfold. First, to quantify where GPT-5 and GPT-4o improve upon GPT-4 in biomedical text-only and multimodal reasoning settings. Second, to assess how far general-purpose LLMs have progressed toward task-specific state-of-the-art systems across extraction, generation, and reasoning tasks. Third, to profile the trade-offs between accuracy, cost, and efficiency in practical deployment scenarios. Finally, to provide empirical guidance for model selection and benchmarking standards in biomedical natural language understanding.

# Methods

## Tasks, datasets, and evaluation metrics

We tested LLMs on two sets of benchmarks that complement each other well. The first set included five core BioNLP tasks that tested basic skills in biomedical text mining. The second set included nine biomedical QA datasets that tested higher-level reasoning and factual consistency. Table 1 provides a summary of all the datasets, task definitions, and evaluation metrics associated with them.

**Table 1. Evaluation datasets and metrics.**

| Datasets | Description | Training | Validation | Testing | Metrics |
|---|---|---|---|---|---|
| **Named Entity Recognition** | | | | | |
| BC5CDR-chemical[22] | PubMed articles annotated with mentions of chemicals and drugs. | 4,560 | 4,581 | 4,797 | Entity-level F1 |
| NCBI-disease[23] | PubMed abstracts annotated with disease mentions and mapped to standardized disease concepts. | 5,424 | 923 | 940 | Entity-level F1 |
| **Relation Extraction** | | | | | |
| ChemProt[24] | PubMed abstracts annotated with chemical–protein interaction relations across multiple categories. | 19,460 | 11,820 | 16,943 | Macro-F1 |
| DDI2013[25] | DrugBank and Medline texts annotated for drug–drug interaction types. | 18,779 | 7,244 | 5,761 | Macro-F1 |
| **Multi-label Document Classification** | | | | | |
| HoC[26] | Biomedical abstracts annotated with one or more of ten cancer hallmark categories. | 1,108 | 157 | 315 | Macro-F1 |
| LitCovid[27] | A curated literature hub to track up-to-date multi-label topic classification corpus of COVID-19 articles in PubMed. | 24,960 | 6,239 | 2,500 | Macro-F1 |
| **Text Summarization** | | | | | |
| PubMed Summarization[28] | Full biomedical articles paired with their expert-written abstracts, supporting single-document summarization. | 117,108 | 6,631 | 6,658 | Rouge-L |
| MS^2[29] | Collections of primary research articles aligned with systematic review summaries, requiring synthesis across documents. | 14,188 | 2,021 | - | Rouge-L |
| **Text Simplification** | | | | | |
| Cochrane PLS[30] | Systematic review abstracts paired with corresponding plain-language summaries written for lay readers. | 3,568 | 411 | 480 | Rouge-L |
| PLOS Simplification[31] | Scientific abstracts from PLOS journals paired with author-provided lay summaries. | 26,124 | 1,000 | 1,000 | Rouge-L |
| **Knowledge QA** | | | | | |
| MedQA (USMLE)[32] | The canonical medical licensing exam QA is widely used for LLM medical knowledge and reasoning. | 10,177 | 1,271 | 1,271 | Accuracy |
| MedMCQA[19] | A large-scale Multiple-Choice QA dataset sourced from real-world medical entrance exams, designed to test LLM medical knowledge and reasoning. | 182,822 | 6,150 | 4,183 | Accuracy |
| PubMedQA[33] | Biomedical research questions paired with PubMed abstracts. | 190,142 | 21,127 | 500 | Accuracy |
| MedXpertQA[34] (text) | A comprehensive benchmark featuring both text and multimodal subsets, designed to evaluate LLMs' expert-level medical knowledge and advanced clinical reasoning. | - | 5 | 2,450 | Accuracy |
| **Clinical reasoning QA** | | | | | |
| DiagnosisArena[35] | A dataset derived from clinical case reports in top-tier medical journals is designed to rigorously assess LLMs' professional-level diagnostic reasoning capabilities in complex clinical scenarios. | - | - | 915 | Accuracy |
| MedcaseReasoning[36] | The dataset containing diagnostic QA cases paired with detailed clinician-authored reasoning is designed to evaluate and improve LLMs' ability to align with professional diagnostic reasoning processes. | 13,092 | 500 | 897 | Accuracy |
| **Multimodal QA** | | | | | |
| VQA-RAD[20] | A dataset of clinically generated visual QA about radiology images. | 1,793 | - | 451 | Accuracy |
| PMC-VQA[37] | A large-scale medical Visual QA dataset featuring QA pairs derived from images across various modalities and diseases from PubMed Central. | 176,948 | 5,000 | - | Accuracy |
| SLAKE[21] | A large dataset for medical visual QA, featuring comprehensive semantic labels annotated by experienced physicians and a new structural medical knowledge base. | 9,849 | 2,109 | 2,070 | Accuracy |

NER. Identify biomedical entities (with specific entity types) mentioned in text. We used two well-known NER datasets: BC5CDR-Chemical[22], which contains chemical/drug mentions in PubMed articles, and NCBI-Disease[23], consisting of disease mentions in PubMed abstracts. Models must output the exact span of each entity mentioned. Performance is evaluated by exact match F1-score (the harmonic mean of precision and recall for predicted entity spans). This strict F1 metric requires the model to produce entity names exactly as they appear in text, matching both boundaries and spelling.

RE. Determine relationships between pairs of entities in a sentence or passage. We evaluated two standard RE tasks: ChemProt[24] (chemical–protein interactions in literature) and DDI2013[25] (drug–drug interactions). Each dataset provides texts with marked entities, and the task is to classify the relationship type (from a fixed set of relation labels) that holds between the entities (or “None” if no relation). Following prior work, we use macro-averaged F1 as the metric for analysis.

Multi-label Document Classification. Assign one or more labels (from a fixed set) to a document, where multiple labels can apply simultaneously. We used the Hallmarks of Cancer (HoC) dataset[26], which has 10 binary labels for cancer biology topics assigned to PubMed abstracts, and LitCovid[27], which assigns up to 7 topic labels (e.g., Treatment, Diagnosis) to COVID-19 research articles. Unlike single-label classification, models may output several labels per document. We evaluate with the macro-averaged F1-score (averaging F1 across all labels).

Text Summarization. The summarization task assesses a model's ability to generate concise yet informative summaries of biomedical scientific texts. We included two benchmark datasets. The first, PubMed Summarization[28], is a single-document task in which the input is the full text of a PubMed article and the target is its gold-standard abstract. The second dataset, MS^2[29] (Multi-Document Summarization of Medical Studies), provides a collection of primary research articles cited within a systematic review; the model must generate a summary that aligns with the systematic review's abstract. This setting is more challenging as it requires synthesizing information across multiple documents and reconciling potentially conflicting evidence. For both datasets, we use ROUGE-L F1 as an evaluation metric, consistent with prior work.

Text Simplification. The simplification task evaluates a model's ability to rewrite biomedical and clinical text into a more accessible form while preserving the original meaning. We considered two datasets. The first, Cochrane Plain Language Summaries (PLS)[30], pairs systematic review abstracts with corresponding plain-language summaries written by experts. Models are required to generate simplified versions of technical content that retain accuracy and fidelity. The second dataset, PLOS Simplification, is constructed from articles in the Public Library of Science (PLOS) journals[31], where scientific abstracts are aligned with their author-written lay summaries. Compared to Cochrane PLS, this dataset often involves more diverse writing styles and varying levels of simplification, making it a complementary evaluation resource. For both datasets, we adopt ROUGE-L F1 as the primary evaluation metric, consistent with previous studies.

To extend beyond core language-processing capabilities, we evaluated models on nine biomedical QA datasets that collectively test factual grounding, clinical reasoning, and multimodal comprehension. MedQA[32] and MedMCQA[19] comprise extensive collections of multiple-choice questions derived from the United States Medical Licensing Examination (USMLE) and other medical board exams. These tasks assess a model's ability to think logically about clinical situations, select the most likely answer from a list of incorrect options, and integrate medical knowledge with logical reasoning. MedXpertQA[34] focuses on diagnostic and treatment questions that have been carefully selected by experts. It focuses on multi-step reasoning and understanding the context. PubMedQA[33] is an example of evidence-based QA. It requires the model to answer 'yes', 'no', or 'maybe' based on a PubMed abstract as supporting evidence. This test assesses how well the facts align and how effectively the model handles uncertainty. DiagnosisArena[35] and MedCaseReasoning[36] concentrate on clinical-reasoning tasks based on authentic case reports, prioritizing differential diagnosis, temporal dynamics, and therapeutic decision-making over the simple retrieval of biomedical facts. Lastly, VQA-RAD[20] and SLAKE[21] incorporate multimodal QA by combining medical images (such as radiographs) with text questions. This test cross-modal understanding and visual grounding. Across all QA datasets, models were prompted in a standardized format and evaluated using accuracy as the primary metric.

For the core BioNLP tasks, we further compared GPT-5 and GPT-4o against previously published state-of-the-art (SOTA) task-specific baselines, following exactly the same data partitions, evaluation metrics, and experimental protocol established by Chen et al.[16]. These reference systems—typically supervised transformer architectures such as BioBERT[10], PubMedBERT[38], and other domain-adapted models—represent the strongest fine-tuned results available for NER, RE, document classification, summarization, and simplification benchmarks. Because our benchmark replicated the identical dataset splits and evaluation pipeline used in that prior work, the published SOTA results could be directly reused for comparison without retraining. While these specialized models deliver high accuracy under task-specific supervision, their reliance on extensive labeled data and narrow transferability across task types underscores the complementary value of assessing general-purpose LLMs within a unified, cross-task framework.

## Models Evaluated and Experimental Setup

This study systematically evaluated the performance of GPT-5 and GPT-4o across all benchmark tasks utilizing a standardized and replicable protocol. The official OpenAI APIs were used to make predictions directly, which ensured they were the same as those made in production-grade inference settings. There were two operational configurations for GPT-5 and GPT-4o. The standard endpoint was used for QA and reasoning-intensive tasks, while the batch API was applied to core BioNLP tasks to optimize cost and latency. This distinction aligns with practical deployment scenarios and provides a basis for analyzing the performance–efficiency trade-offs reported in the cost evaluation section. To ensure consistency and traceability, we systematically recorded token usage and inference cost for every model–task pair. For each generation, the total number of input and output tokens was logged through the API metadata, allowing precise calculation of token-normalized cost according to the official OpenAI pricing for both standard and batch endpoints. These statistics were aggregated at the dataset and task levels to quantify efficiency alongside accuracy. In addition, latency—defined as the total time between prompt submission and response completion—was automatically measured for each batch.

As historical baselines, results for GPT-4 and GPT-3.5 were obtained from the prior benchmark study by Chen et al.[16]. Those evaluations utilized the same prompt templates, decoding parameters, and dataset partitions, which enabled direct model comparison without rerunning the old systems. This alignment enables the conduct of meaningful, long-term evaluations of how models have improved across different generations of LLMs. It also ensures that any improvements are due to architectural and alignment changes, not changes in protocol.

We tested all the models in three standard ways: zero-shot, one-shot, and five-shot. In the zero-shot condition, models received only the task description and input text, testing their generalization without exposure to examples. In the one-shot and five-shot settings, prompts additionally included one or five representative input–output exemplars drawn from the training split of each dataset. The same exemplars were used across all models and kept fixed throughout the study to eliminate bias from exemplar selection. This design isolates performance differences attributable to model reasoning and alignment rather than prompt variability.

For selected QA datasets that contained explicit subtype annotations (e.g., clinical reasoning type, question modality, or content domain), we further performed subject-level error and accuracy analyses. Each question instance was categorized by its underlying subtype (e.g., anatomy, diagnosis, pharmacology, image-based reasoning), enabling fine-grained comparison of GPT-5 and GPT-4o across reasoning dimensions. These analyses offered further clarity regarding the areas of most significant model enhancement and the biomedical knowledge categories that continued to pose difficulties.

The experimental setup overall replicates and builds upon the protocol set up in the last benchmark, ensuring that task definitions, decoding parameters, and evaluation metrics are all consistent. This unified framework establishes a controlled environment for comparing large language models across various biomedical NLP tasks. It also systematically combines measurements of accuracy, latency, and cost so that cross-model evaluation is clear and repeatable.

## Results

### Overall performance across core BioNLP tasks

The updated benchmark indicates that improvements are clear at the frontier. Figure 1 shows that GPT-5 always got the best overall scores when prompted with zero, one, or five shots. In macro-average terms across all 10 datasets, GPT-5 improved from 0.477 in the zero-shot condition to 0.502 in the five-shot condition. By comparison, GPT-4 increased from 0.413 to 0.455, while GPT-4o rose only slightly from 0.445 to 0.461. These results demonstrate that few-shot demonstrations continue to provide modest benefits, but the marginal gains diminish in larger models. Relative to previously reported task-specific SOTA systems, GPT-5 narrows the performance gap, although the supervised SOTA models still maintain an advantage.

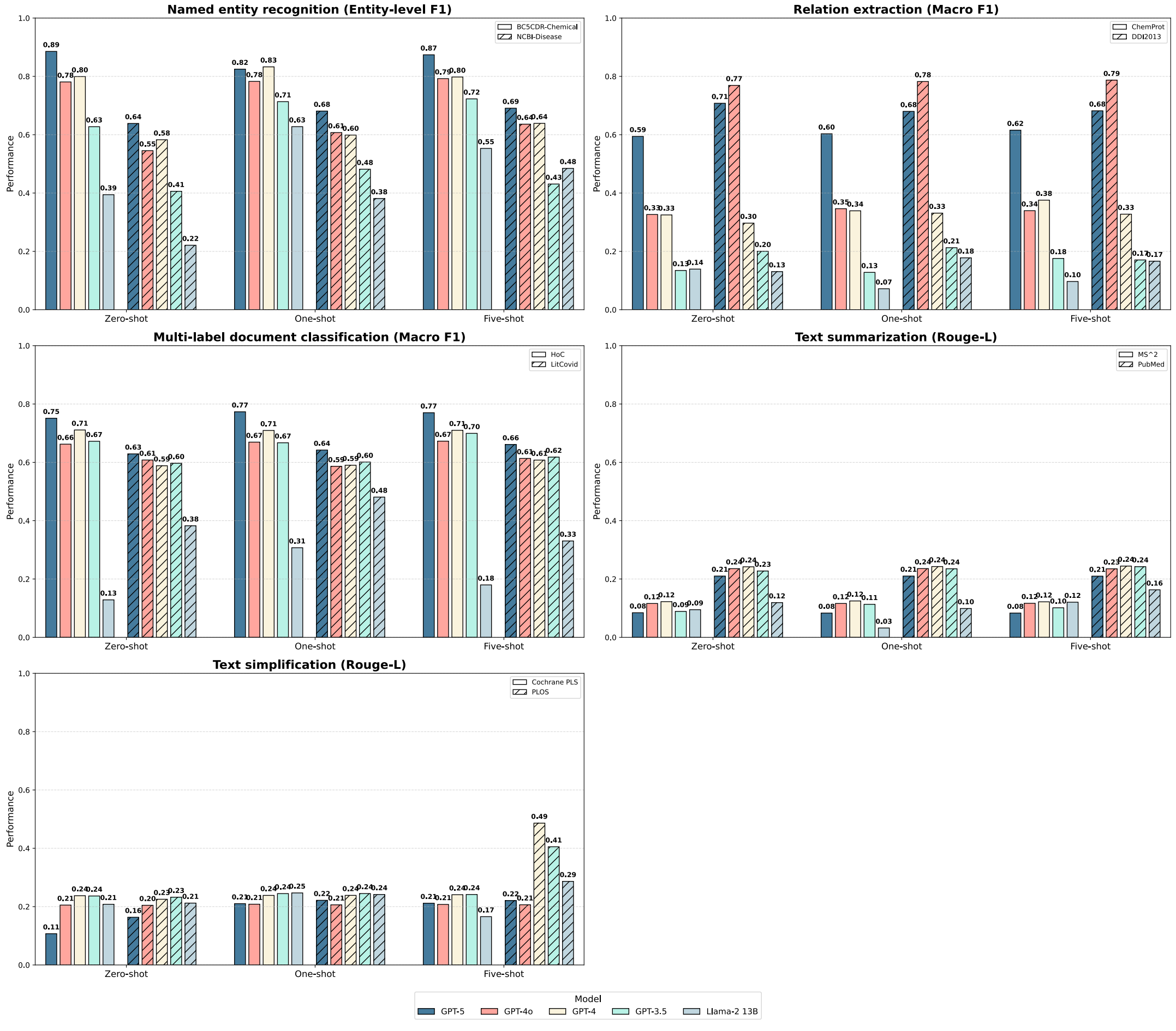

**Figure 1. Quantitative evaluations of the LLMs on the five core BioNLP tasks under zero/few-shot**

On the BC5CDR-Chemical in NER task, GPT-5 achieved 0.886 F1 in zero-shot and 0.874 in five-shot, within approximately six to eight points of SOTA (0.950). On NCBI-Disease, however, GPT-5 reached only 0.691 in five-shot, which is still more than twenty points lower than SOTA (0.909). These results indicate that GPT-5 substantially improves chemical recognition but struggles with disease entities. In the RE task, the results varied substantially by dataset. On ChemProt, GPT-5 reached 0.616 F1 in five-shot, reducing the gap to SOTA to approximately twelve points. In contrast, on DDI2013, GPT-4o achieved 0.787 in five-shot, essentially matching SOTA (0.792) and outperforming GPT-5. This divergence suggests that task schema and prompt adherence strongly influence LLM performance. For the HoC in the Multi-label document classification task, GPT-5 achieved 0.770 macro-F1 in five-shot, while for LitCovid, it achieved 0.661. Although these results outperform GPT-4 and GPT-4o, they remain between twelve and twenty-three points lower than SOTA.

On PubMed single-document summarization, GPT-5 achieved approximately 0.21 ROUGE-L, compared with a SOTA of 0.43. On MS^2 multi-document summarization, GPT-5 remained at approximately 0.08, compared with 0.21 for SOTA. Few-shot prompting didn't provide a benefit in a measurable way. Outputs were generally shorter than reference abstracts, indicating brevity bias and inadequate coverage as constraining factors. In the Text simplification task on Cochrane PLS, GPT-5 went from 0.107 in zero-shot to 0.212 in five-shot, but it was still below SOTA (0.448). The supervised SOTA baseline (0.437) outperforms the PLOS score of 0.486 for GPT-4 with five demonstrations. GPT-5, on the other hand, stayed lower (0.221 in five-shot). These results show that GPT-4 does better with style priming based on examples.

## Expansion to biomedical QA tasks

We added nine additional biomedical QA benchmarks to the five main BioNLP tasks to enhance the thoroughness of the evaluation (the performance as shown in Figure 2). These benchmarks tested knowledge retrieval, clinical reasoning, and multimodal understanding. In all of these datasets, GPT-5 showed a steady and big improvement over GPT-4o when prompted in the same way with zero, one, and five shots. GPT-5 achieved an average accuracy of 0.762, 0.737, and 0.768 across all datasets, while GPT-4o achieved an average accuracy of 0.665, 0.652, and 0.657. The five-shot comparison, which best approximates practical in-context use, yielded an overall margin of +0.111 in favor of GPT-5.

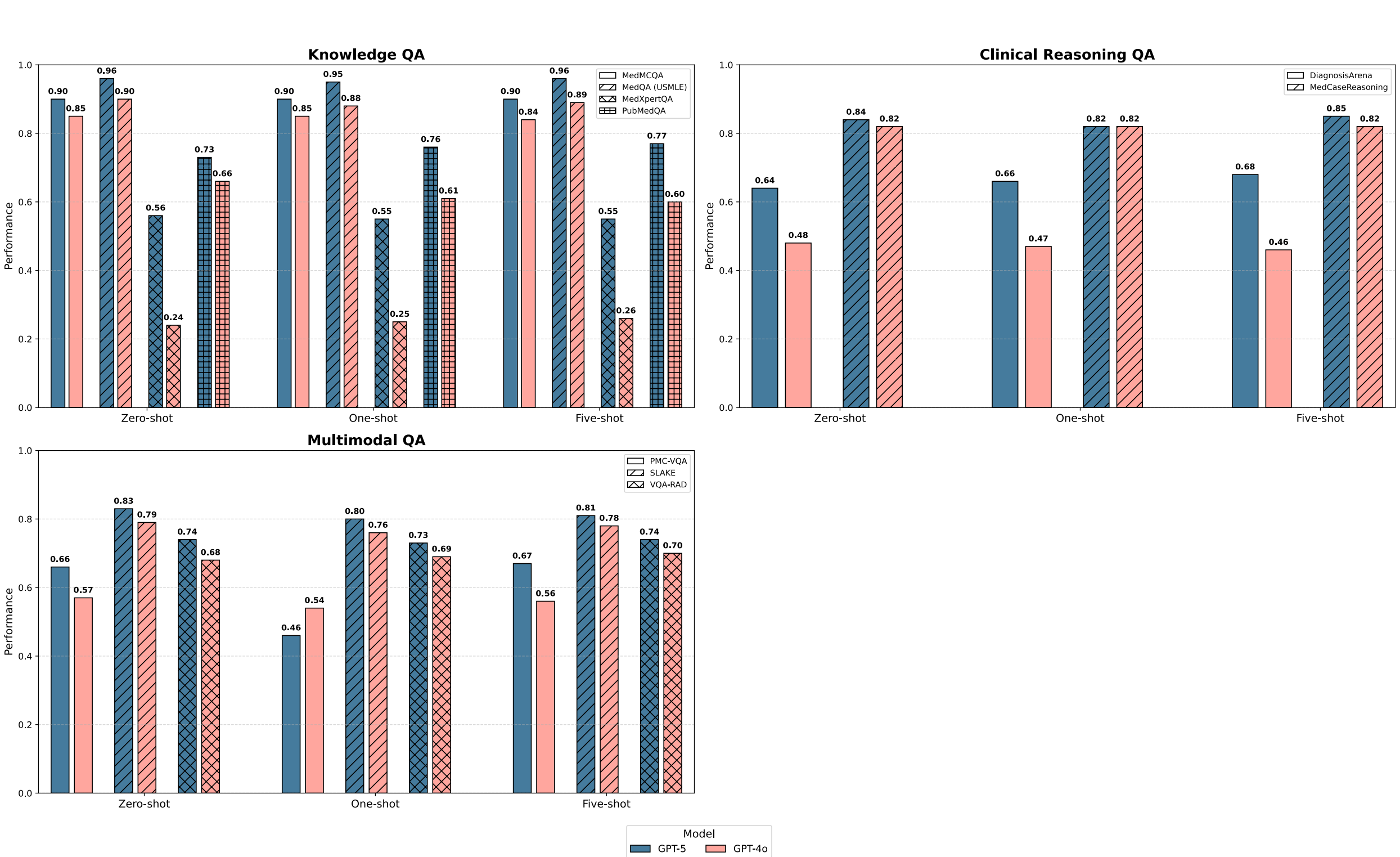

**Figure 2 Quantitative evaluations of the GPT-5 and GPT-4o across biomedical QA under zero/few-shot**

Performance gains were not confined to a single task category. In the knowledge-oriented benchmarks, GPT-5 markedly outperformed GPT-4o, reaching 0.956 on MedQA (USMLE) and 0.900 on MedMCQA, compared with 0.888 and 0.845, respectively. In clinical reasoning datasets, the differences were even bigger. For example, GPT-5 scored +0.288 higher than GPT-4o on MedXpertQA (0.548 vs. 0.260) and +0.221 higher on DiagnosisArena (0.681 vs. 0.460). In multimodal QA tasks, GPT-5 also did better: it got 0.742, 0.665, and 0.808 in VQA-RAD, PMC-VQA, and SLAKE, while GPT-4o got 0.705, 0.558, and 0.776. When divided by type, GPT-5's average five-shot accuracy was 0.793 for knowledge QA, 0.764 for clinical reasoning, and 0.738 for multimodal QA. This was better than GPT-4o by +0.14, +0.13, and +0.06, in that order. These results demonstrate that GPT-5 is as reliable in biomedical QA as specialized systems, particularly when complex reasoning is required.

A closer examination of the question subtypes helps to illuminate where these performance differences originate (Figure 3). Within MedXpertQA, the largest improvements were observed for Diagnosis (+0.36), Reasoning (+0.34), and Treatment (+0.30) items, with additional though smaller gains in Basic Science (+0.27) and Understanding (+0.26). These patterns indicate that GPT-5's advantage is most evident in tasks requiring multi-step reasoning and the synthesis of distributed biomedical evidence to form coherent clinical judgments. A discipline-level breakdown of MedMCQA results shows that GPT-5 consistently outperformed GPT-4o across nearly all medical specialties. The most pronounced increases appeared in Orthopedics (+0.09), Otolaryngology (ENT) (+0.08), and Anatomy (+0.06), followed by moderate yet steady gains in Ophthalmology (+0.05) and other foundational domains such as Pharmacology, Microbiology, and Pathology (typically +0.03 to +0.04). This uniformity across diverse subjects suggests a general improvement in biomedical representation rather than overfitting to any single knowledge area, consistent with GPT-5's stronger handling of terminology variation and cross-disciplinary inference. In the multimodal context of SLAKE, where each question pairs textual input with radiological imagery, GPT-5's benefits were most visible for attributes demanding spatial reasoning or contextual interpretation. Accuracy gains were striking for Position (+0.17), Modality (+0.09), and Abnormality (+0.07), whereas categories relying on surface-level visual cues—such as Color and Organ—showed only marginal change. Across imaging types, GPT-5 reached its best overall accuracy on MRI (0.909 vs. 0.818, Δ +0.09), followed by X-ray (Δ +0.06) and CT (Δ +0.02), reflecting a more refined semantic alignment between visual data and textual reasoning.

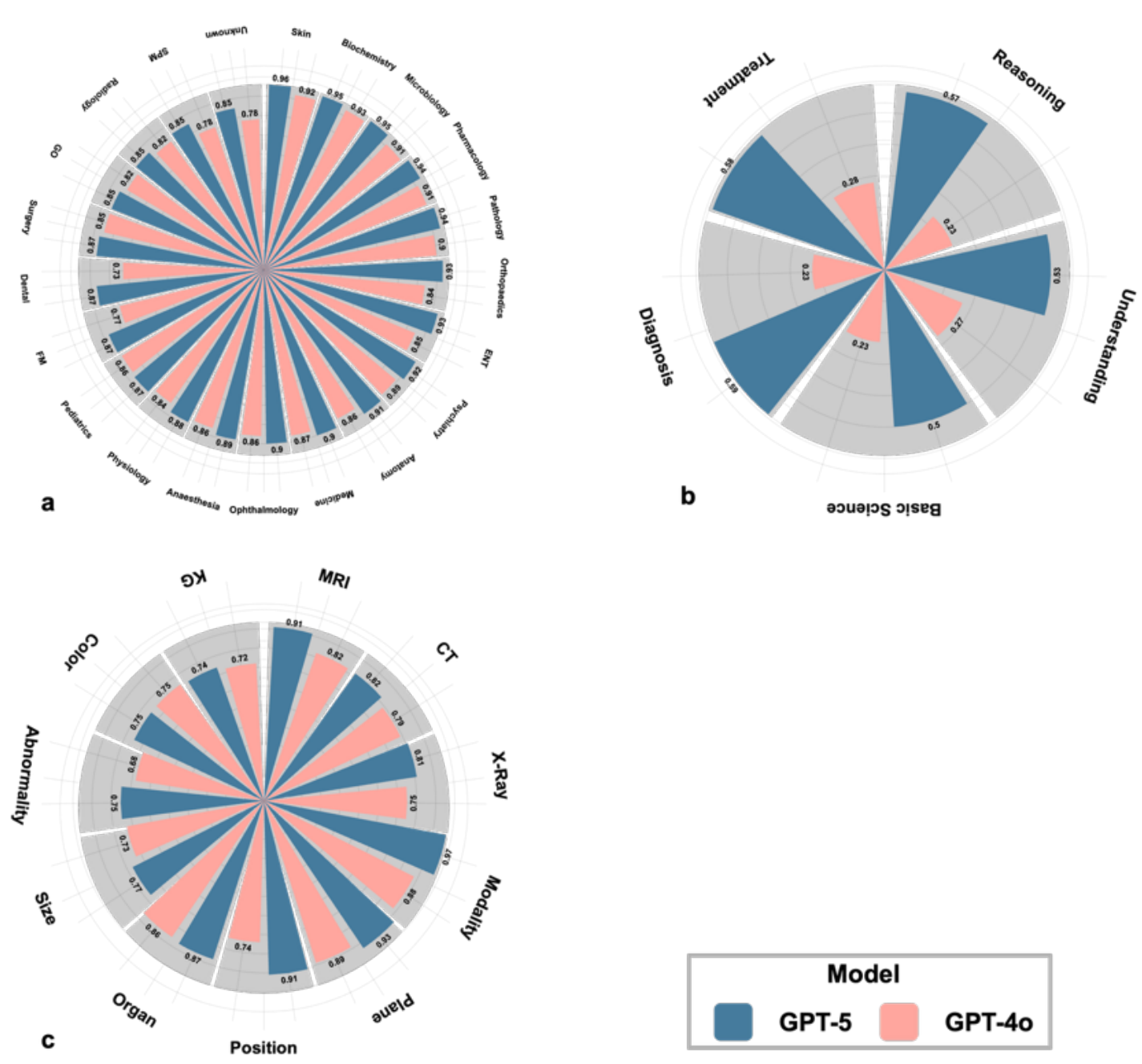

**Figure 3 Fine-grained analysis of GPT-5 and GPT-4o performance across biomedical QA subtypes. (a) MedMCQA. (b) MedXpertQA. (c) SLAKE**

## Cost and efficiency analysis

To complement accuracy-based benchmarking, we examined the computational and economic efficiency of GPT-5 and GPT-4o under the official OpenAI pricing released in October 2025. All five core BioNLP tasks were executed through the batch API, while the expanded biomedical QA benchmarks used the standard API. This division reflects realistic deployment settings in which high-throughput text mining is performed in batch mode and interactive question answering requires standard, low-latency inference.

For the main BioNLP tasks, GPT-5 took in about the same number of input tokens as GPT-4o, but its outputs were always longer and more detailed. This pattern indicates that GPT-5 tends to provide comprehensive answers that align with the situation rather than brief summaries. The output length did go up, but using the batch API saved about 45 to 50 percent compared to GPT-4o, which costs $1.25 and $5.00. The batch API costs $5.00 for every million output tokens and $0.625 for every million input tokens. For representative datasets such as BC5CDR-Chemical and DDI2013, the total expenditure per 1,000 processed instances declined from about $0.0036 with GPT-4o to $0.0019 with GPT-5, assuming equivalent token throughput. Average latency remained in a similar range, between 300 and 450 milliseconds, suggesting that GPT-5's longer responses did not introduce a noticeable delay in practical inference.

In the QA benchmarks, which were executed using the standard API, the unit prices were substantially higher because of the $10-per-million-token rate applied to outputs. Even under these conditions, GPT-5 retained a clear economic advantage. Its price range ($1.25 for input and $10 for output) was lower than GPT-4o's ($2.50 and $10, respectively), and its token use was more stable across the zero-, one-, and five-shot configurations. After normalization by the actual

token counts summarized in Appendix Table A1, the average cost for processing 100 QA samples was around $0.72 for GPT-5 and $1.06 for GPT-4o, representing an estimated 30 to 40 percent reduction in cost per correct answer. Latency again remained comparable, typically between 300 and 500 milliseconds.

## Discussion

This study aimed to investigate the potential of a cutting-edge general-purpose model (GPT-5) as a practical and empirically validated benchmark for biomedical NLP, providing dependable accuracy and cost-effective inference across various tasks. Our goal with a comprehensive benchmark that includes five core BioNLP tasks and nine extended biomedical QA datasets was not only to compare model performance but also to see how improvements in architecture and alignment affect reasoning robustness, efficiency, and real-world usability. Instead of focusing on overall leaderboard gains, the study aimed to provide evidence that clarifies the appropriate timing and methods for deploying general LLMs in biomedical research and clinical informatics settings. Overall, GPT-5 showed clear advantages over GPT-4o, especially in tasks that required a lot of reasoning and QA, and it had much better cost–performance ratios when using standard API pricing. In summary, these findings suggest that frontier general-purpose models are approaching a stage of maturity where a balanced evaluation of accuracy, interpretability, and efficiency can significantly impact their practical implementation in biomedical NLP pipelines.

In the context of previous research, our findings align with an expanding corpus of literature indicating that meticulously prompted general-purpose LLMs can match or even exceed domain-specific systems in medical reasoning and explanatory writing tasks, such as MedQA and plain-language summarization. A particularly pertinent comparison is the recent assessment of GPT-5 on the MedXpertQA-Text dataset[39], which indicated performance levels largely analogous to ours, despite utilizing an explicit chain-of-thought (CoT) prompting strategy, whereas our experiments depended on simple single-turn prompts. This convergence suggests that GPT-5's baseline reasoning ability may already approach the level previously attainable only through structured CoT scaffolds. From a practical standpoint, these results imply that for many structured biomedical QA tasks, direct prompting can achieve competitive outcomes, particularly when computational efficiency or deployment simplicity are priorities. Nevertheless, the modest yet consistent gains observed under CoT prompting in that study suggest that there may be remaining benefits for tasks that demand extended causal reasoning or diagnostic synthesis. These observations collectively endorse a tiered prompting methodology in biomedical applications. Standard direct prompting may be sufficient for extensive or resource-sensitive pipelines where efficiency and cost are critical. In contrast, CoT-enhanced or multi-step reasoning templates may be designated for high-stakes or analytically complex tasks requiring explicit justification and clarity. In this context, our findings enhance previous research by pinpointing an evolving equilibrium between reasoning depth and operational efficiency in cutting-edge general-purpose LLMs, thereby offering a practical framework for prompt-strategy selection in biomedical NLP applications. In addition, while the proposed tiered prompting framework provides practical guidance for most reasoning and QA scenarios, certain task families remain notably resistant to prompt-based improvements. Consistent with prior evidence, summarization and span-level extraction continue to pose challenges for general LLMs, with recurring issues in content selection, factual coverage, boundary precision, and length control[40–42]. These tasks frequently rely less on abstract reasoning and more on precise linguistic fidelity, potentially limiting the advantages of even well-aligned general models. In these instances, specialized or hybrid systems—integrating retrieval, domain ontologies, or structured supervision—are likely to remain indispensable for attaining the accuracy, consistency, and factual precision required in biomedical text processing.

A deeper comparison with earlier studies clarifies where our conclusions extend and where they depart. First, we corroborate the scaling trend reported for GPT-4-series and clinical-tuned models that larger, better-aligned systems tend to improve on medical QA and explanation[43–47]. Still, we also show that the incrementalism observed near GPT-4 is less evident at GPT-5, which exhibits robust gains across modalities and cognitive skill categories in our extended QA battery. Second, consistent with findings that simple few-shot prompting yields diminishing returns as models become more capable[47,48], we observed modest benefits from additional exemplars for GPT-5/GPT-4o, with the most significant effects confined to stylistically sensitive tasks; conversely, reasoning-heavy QA and multimodal items showed strong performance with minimal exemplars, suggesting that frontier models increasingly rely on intrinsic priors rather than prompt-provided scaffolding. Third, where our results differ from studies reporting limited improvements beyond GPT-4, plausible explanations include differences in prompt discipline, evaluation granularity, and the absence in prior work of cost-normalized metrics that can materially alter recommendations even when raw accuracies are close. Finally, our cost analysis complements accuracy-focused reports by quantifying that GPT-5's batch and standard pricing translate into 30–50% lower effective cost per correct answer compared with GPT-4o in otherwise matched conditions. This consideration directly informs deployment decisions.

The synthesis of these observations supports a recommendation-oriented view. For knowledge-intensive and clinically structured QA, particularly tasks that require multi-step constraint propagation and decision consistency, GPT-5 should be preferred as a general model, as it delivers higher accuracy and lower cost per correct output than GPT-4o under both batch and standard APIs. For span-level extraction and high-stakes summarization, our results reaffirm the value of hybridization: retrieval-augmented prompting, task-specific calibration, and knowledge-graph–grounded verification can remedy failure modes that persist for general models[40–42,49–51]. In settings where regulatory or operational constraints prioritize boundary fidelity, exhaustive recall, and evidence traceability, domain-tuned components remain advisable, either as primary models or as validators layered on top of GPT-5.

Notwithstanding these strengths, several limitations qualify the scope of our conclusions. First, although we expanded the coverage to include nine QA datasets and retained five core BioNLP tasks, automated metrics are still insufficient to accurately measure clinical usefulness, as they fail to fully demonstrate factual faithfulness, calibration, or safety. This might not accurately predict practical gains, especially when it comes to summarization and extraction[40,41]. Second, the proprietary nature of GPT-5 and GPT-4o precludes architectural ablations and dataset transparency, limiting interpretability and replication. Third, while we controlled prompting and decoding, we did not exhaust the design space of structured prompting, tool use, or retrieval-augmented workflows; prior studies show that such strategies can yield additional improvements beyond static few-shot[47–51]. We mitigated these constraints by standardizing prompts, reporting latency alongside token-normalized cost, and analyzing subtype-level performance to localize strengths and weaknesses, but a fuller treatment will require human evaluation and prospective clinical tasks.

These caveats suggest concrete next steps for method and practice. Benchmarks should pair automatic metrics with expert review to assess faithfulness, harmful error modes, and clinical plausibility, particularly for summarization and extraction[40,41]. Where boundary precision or evidence completeness is critical, hybrid pipelines that combine GPT-5 with retrieval augmentation, ontology-aware constraint checking, or knowledge graph verification are likely to be more reliable[49–51]. For QA and decision support, cost-aware deployment should exploit batch processing for high-throughput back-ends and standard endpoints for interactive use, with explicit monitoring of the cost-per-correct-answer rather than raw token counts. Finally, reporting should include token-normalized costs and latency to facilitate end-user budgeting and system sizing, as modest accuracy differences can be outweighed by economic efficiency in practice.

Looking forward, three directions appear most consequential. First, extending evaluation to clinical notes, multilingual corpora, and richer multimodal contexts will better characterize generalization under real-world data distributions. Second, integrating retrieval-augmented reasoning, tool invocation, and knowledge-graph grounding within a single evaluation harness can clarify how much residual error is attributable to missing evidence versus model reasoning, and whether hybridization narrows the remaining gaps in summarization and NER. Third, prospective studies that couple cost-normalized metrics with human-in-the-loop assessment and outcome-relevant endpoints are needed to translate benchmark gains into safe, auditable workflows. Taken together, our results suggest a hybrid future in which GPT-5 functions as a strong general scaffold—cost-efficient and reasoning-capable—while domain-tuned and retrieval-grounded components deliver the precision, faithfulness, and traceability required for high-stakes biomedical applications.

## Conclusion

This study offers a thorough and methodologically sound benchmark of GPT-5 and GPT-4o across five fundamental BioNLP tasks and nine biomedical QA datasets, enhancing accuracy-focused evaluation with token-level cost and latency metrics. GPT-5 shows strong improvements over GPT-4o, especially in reasoning-oriented and multimodal QA. It also closes the gaps on some extraction and classification tasks. Disease NER, multi-label classification, and evidence-dense summarization still work best with domain-tuned or hybrid pipelines. This indicates that there are still issues with boundary fidelity, content selection, and factual accuracy. From a deployment perspective, GPT-5 can achieve competitive accuracy at a lower cost-per-correct-answer with similar latency by utilizing standardized prompting, batch execution for core text-mining workloads, and systematic cost tracking. These results suggest that a tiered prompting strategy is a practical approach to implementing solutions in the real world: direct prompting may be sufficient for large or cost-sensitive applications, while chain-of-thought or multi-step scaffolds are more suitable for high-stakes tasks that require clear reasoning and understanding. In the future, it will be crucial to integrate retrieval-augmented reasoning, ontology-aware verification, and knowledge-graph grounding to further enhance factual rigor and traceability.